%% file: main.tex
\documentclass{article}
\usepackage[nonatbib, preprint]{neurips_2023}

\usepackage{graphicx}
\usepackage{latexsym}
\usepackage{subcaption}
\usepackage{amsthm}
\usepackage{amsmath}
\usepackage{algorithm}
\usepackage[noend]{algpseudocode}
\usepackage{multirow}
\usepackage{utfsym}
\usepackage{float}
\usepackage{wrapfig}
\usepackage[colorlinks=true, allcolors=blue]{hyperref}
\usepackage[utf8]{inputenc} 
\usepackage[T1]{fontenc}    
\usepackage{url}            
\usepackage{booktabs}       
\usepackage{amsfonts}       
\usepackage{nicefrac}       
\usepackage{microtype}      
\usepackage{xcolor}         

\title{pFedSim: Similarity-Aware Model Aggregation Towards Personalized Federated Learning}

\author{%
  Jiahao~Tan \\
  Shenzhen University, P.R. China \\
  \texttt{2100271083@email.szu.edu.cn} \\
  \And
  Yipeng~Zhou \\
  Macquarie University, Australia \\
  \texttt{yipeng.zhou@mq.edu.au}
  \And Gang~Liu\thanks{Corresponding Author} \\
  Shenzhen University, P.R. China \\
  \texttt{gliu@szu.edu.cn}
  \And
  Jessie~Hui~Wang \\
  Tsinghua University, P.R. China \\
  \texttt{jessiewang@tsinghua.edu.cn}
  \And
  Shui~Yu \\
  University of Technology Sydney, Australia \\
  \texttt{Shui.Yu@uts.edu.au}
}

\begin{document}

\maketitle

\begin{abstract}
  \input{abstract.tex}
\end{abstract}

\section{Introduction}
\input{introduction.tex}

\section{Related Works}

\input{related_works.tex}

\section{Preliminary}
\input{preliminary.tex}

\section{pFL Algorithm Design}
\input{method.tex}

\section{Experiment}
\input{experiment.tex}

\section{Conclusion}
\input{conclusion.tex}

\bibliographystyle{IEEEtran}
\bibliography{references}

\newpage
\appendix
\input{appendix.tex}

\end{document}

%% file: abstract.tex
The federated learning (FL) paradigm emerges to preserve data privacy during model training by only exposing clients' model parameters rather than original data.  One of the biggest challenges in FL lies in the non-IID (not identical and independently distributed) data  (a.k.a., data heterogeneity) distributed on clients. To address this challenge, various personalized FL (pFL) methods are proposed such as similarity-based aggregation and model decoupling. The former one aggregates models from clients of a similar data distribution. The later one decouples a neural network (NN) model into a feature extractor and a classifier. Personalization is captured by classifiers which are obtained by local training.  To advance pFL, we propose a novel $\texttt{pFedSim}$ (pFL based on model similarity) algorithm in this work by combining these two kinds of methods. More specifically, we decouple a NN model into a personalized feature extractor, obtained by aggregating models from similar clients,  and a classifier, which is obtained by local training and used to estimate client similarity.  Compared with the state-of-the-art baselines, the advantages of $\texttt{pFedSim}$ include: 1) significantly improved model accuracy; 2) low communication and computation overhead; 3) a low risk of privacy leakage; 4) no requirement for any external public information. To demonstrate the superiority of $\texttt{pFedSim}$, extensive experiments are conducted on real datasets. The  results validate the superb performance of our algorithm which can significantly outperform baselines under various heterogeneous data settings.

%% file: introduction.tex
Federated Learning (FL) is an emerging paradigm that can preserve data privacy while training machine learning models. 
In FL \cite{fedavg}, a parameter server (PS), \emph{e.g.}, a cloud server, is deployed to coordinate the training process over multiple global iterations. In each global iteration, the PS   communicates with participating FL clients owning original private data for exchanging model parameters.  A significant challenge in FL lies in the statistical heterogeneity of data owned by different clients (\emph{i.e.}, non-IID data) because the data generated by different clients obeys distinct distributions \cite{fedavg}. Consequently, a single global model trained by FL fails to fit heterogeneous data on individual clients very well. In extreme cases, the data heterogeneity can severely lower model utility, slow down the FL convergence, and even make FL diverge \cite{fl_with_noniid_data}.  

To tackle the challenge of non-IID data on FL clients, personalized FL (pFL) is proposed with the principle to customize models for individual clients. Until now, similarity-based aggregation and model decoupling are two most widely studied approaches to achieve pFL. The principle of the former approach is to aggregate clients of a similar data distribution so that a personalized model can be produced to fit local data \cite{fedprox, scaffold, moon, feddyn, fedavgm}.  
However, original data is invisible in FL making similarity estimation difficult. Most existing works require FL clients to expose additional information such as statistical knowledge of data label distribution for similarity estimation \cite{fedproto, fedap}. Whereas, exposing additional information may incur heavy communication/computation overhead and make clients suffer from privacy leakage \cite{flcl, dlg}. For example, $\texttt{FedAP}$ \cite{fedap} required FL clients to expose statistical knowledge of their private data  to estimate client similarity. The Wasserstein distance of normal distributions generated by running statistics of two arbitrary clients' local batch normalization layers is used to measure similarity, which is then used to guide model aggregation under severe non-IID data settings. 
To avoid exposing additional information to the federated learning (FL) server, FL clients in \texttt{FedFomo} \cite{fedfomo} receive additional models belonging to their neighboring clients from the PS to gather knowledge and guide personalized aggregation. However, this can significantly aggravate the communication burden of FL clients. Another approach is to decouple a neural network (NN) model into a feature extractor and a classifier. Previous studies \cite{fedper, fedrep} suggest that the final fully connected layer in CNN models, such as \cite{lenet5, mobilenetv2}, should be included in the classifier part, while other layers should be included in the feature extractor part. The classifier is mainly updated by local training to achieve personalized performance, while the feature extractor is trained across all FL clients to fully utilize all data in the system.

Different from existing works, we propose a novel \texttt{pFedSim} (pFL with model similarity) algorithm by combining similarity-based aggregation and model decoupling methods. More specifically, \texttt{pFedSim} decouples a neural network model as a personalized feature extractor and a classifier. Client similarity is measured by the distance of classifiers, and personalized feature extractors are obtained by aggregating over similar clients. 
Considering that model parameters are randomly initialized, we design \texttt{pFedSim} with two phases: generalization and personalization. In the generalization phase, traditional FL algorithms, \emph{e.g.}, \texttt{FedAvg}, is executed for model training. The personalization phase is an iterative learning process with two distinct operations in each global iteration, which includes:  1) Refining feature extractors and classifiers by local training.   2) Similarity (measured based on the classifier distance) based feature extractor aggregation to fully utilize data across similar clients.
Compared with existing model decoupling methods, \texttt{pFedSim} can significantly improve model accuracy by personalizing the feature extractor part. 
Compared with existing similarity-based methods, \texttt{pFedSim} can more accurately capture client similarity based on the classifier distance. Meanwhile, \texttt{pFedSim} averts heavy communication/computation overhead and privacy leakage risks because no additional information other than model parameters is exposed. Our empirical study by using real datasets, \emph{i.e.}, CIFAR-10, CINIC-10, Tiny-ImageNet and EMNIST, demonstrates that \texttt{pFedSim} significantly outperforms the state-of-the-art baselines under various non-IID data settings. The experiment results demonstrate that \texttt{pFedSim} can improve model accuracy by 2.96\% on average.

To have a holistic overview of the advantages of \texttt{pFedSim}, we qualitatively compare it with the state-of-the-art baselines in Table~\ref{tb:req} from four perspectives: communication load, privacy leakage risk, computation load and  requirement for external data (which is usually unavailable in practice).
Through the comparison, it is worth noting that \texttt{pFedSim} is the only one without any shortcomings listed in Table~\ref{tb:req} because our design is only based on exposed model parameters.

%% file: related_works.tex
In this section, we overview prior related works and discuss our contribution in comparison to them. 

\noindent{\bf Traditional FL} Traditional FL aims to train a single global model to fit  data distributed on all clients. 
The first traditional model training algorithm in FL is \texttt{FedAvg} \cite{fedavg}, which improves training efficiency by updating the model locally over multiple rounds to reduce the  communication frequency. However, \texttt{FedAvg} fails to consider the non-IID property of data in FL, and thus its performance is inferior under heterogeneous data settings. 

\begin{table*}
    \centering
    \resizebox{\textwidth}{!}{
        \begin{tabular}{l|cccc|c}
            Weakness                 & \texttt{FedFomo} \cite{fedfomo} & \texttt{FedAP} \cite{fedap}    & \texttt{FedGen} \cite{fedgen}  & \texttt{FedMD} \cite{fedmd}    & \texttt{pFedSim} (Ours)        \\ 
            \midrule
            Heavy Communication Load & $\checkmark$                    & $\scalebox{0.75}{\usym{2613}}$ & $\checkmark$                   & $\checkmark$                   & $\scalebox{0.75}{\usym{2613}}$ \\
            High Privacy Risk        & $\scalebox{0.75}{\usym{2613}}$  & $\checkmark$                   & $\checkmark$                   & $\scalebox{0.75}{\usym{2613}}$ & $\scalebox{0.75}{\usym{2613}}$ \\
            Need Public Data         & $\scalebox{0.75}{\usym{2613}}$  & $\scalebox{0.75}{\usym{2613}}$ & $\scalebox{0.75}{\usym{2613}}$ & $\checkmark$                   & $\scalebox{0.75}{\usym{2613}}$ \\
            Heavy Computation Load   & 
            $\checkmark$             & $\scalebox{0.75}{\usym{2613}}$  & \checkmark                     & $\scalebox{0.75}{\usym{2613}}$ & $\scalebox{0.75}{\usym{2613}}$                                  \\
        \end{tabular}
    }
    \caption{Comparison of shortcomings between \texttt{pFedSim} and different pFL algorithms.}
    \label{tb:req}
    \vspace{-4mm}
\end{table*}

To address the challenge of data heterogeneity, various variants of \texttt{FedAvg} were proposed.  For examples, \cite{fedprox} and \cite{feddyn} introduced a proximal term to clients' optimization objectives so as to normalize their local model parameters and prevent over-divergence from the global model. \cite{fedavgm} added the momentum in model aggregation to mitigate harmful oscillations, which can achieve a faster convergence rate, and hence improve model utility. \cite{fedlc} and \cite{fedalign} changed the empirical loss function, such as cross-entropy loss, to advanced loss functions so as to improve model learning performance. \cite{moon} combined model-based contrastive learning with FL, bridging the gap between representations produced by the global model and the current local model to limit parameter divergence. Despite the progress made by these works in making a single global model more robust in non-IID data scenarios, they overlooked the difference between the global optimization objective and individual clients' diverse local objectives. Thus, their performance is still unsatisfactory. 

\noindent{\bf Personalized FL} Later on, more sophisticated personalized FL was proposed that can optimize the model of FL for each individual client through collaborating with other clients of a similar data distribution. pFL is an effective approach to overcoming the data heterogeneity issue in FL. Existing works have proposed different ways to achieve pFL. \cite{perfedavg, per_fedavg, ditto} were designed based on the similarity between the optimization objectives of various clients in FL, and used meta-learning approaches \cite{maml, reptile} for rapid adaptation of the global model to the local data distribution after fine-tuning at the client side. \cite{mocha} combined multi-task learning with FL that considers each client as a learning task in order to transfer federation knowledge from the task relationship matrix. \cite{pfedhn, pfedla} showed the effectiveness to produce exclusive model parameters and aggregation weights for individual clients by making use of hypernetwork. \cite{fedmd, ktpfl, fedgen} combined the knowledge distillation with FL, which leverage the global knowledge to enhance the generalization of each client model. Nevertheless, these works require the availability of a public dataset which is not always valid in FL scenarios. \cite{apfl, l2gd} mixed the global model with local models to produce a personalized model for each client.

Transmitting additional client information and model decoupling are two commonly used method to boost the performance of pFL. \texttt{FedAP} \cite{fedap} pre-trained  models  on a portion of the training data, and transmitted additional data features to the PS,  which is not easy for implementation in practice. \texttt{FedFomo} \cite{fedfomo} set a similarity matrix to store all client models on the server like \cite{fedap}, but offloaded the model aggregation to the client side.  Based on the similarity matrix, multiple models are distributed to each single client, which will be evaluated by each client based on a local validation set to determine their weights for subsequent model aggregation. However, distributing multiple models to each client can considerably surge the communication cost.  On the contrary, model decoupling achieves pFL with a much lower cost. Decoupling the feature extractor and classifier from the FL model can achieve personalization, which has been explored by \cite{fedper, fedrecon, fedrep, fedbabu, fedrod}. \cite{ccvr} has demonstrated that FL model performance will deteriorate if the classifier is involved into global model aggregation, given  non-IID  data. Therefore, to achieve personalization, the classifier should be retained locally without aggregating with classifiers of other clients. \cite{fedrod} designed dual classifiers for pFL, \emph{i.e.}, the personal one and the generic one, for personalization and generalization purposes, respectively. We design \texttt{pFedSim} by combining similarity based pFL and model decoupling based pFL to exert their strengths without incurring their weaknesses. On the one hand, we advance model decoupling by personalize feature extractors based on classifier-based similarity. On the other hand, \texttt{pFedSim} outperforms existing similarity based pFL methods because \texttt{pFedSim} clients  never exposes overhead information except model parameters for similarity estimation.

%% file: preliminary.tex
We introduce preliminary knowledge of FL and pFL in this section. 

\noindent{\bf FL Procedure}
Let $\mathcal{C} = \{1, \dots, n\}$ denote the set of clients  in FL with total $n$ clients.  The $i$-th client owns the private dataset $\mathcal{D}_i \in \mathcal{X}_i \times \mathcal{Y}_i$, where $\mathcal{X}_i$ and $\mathcal{Y}_i$ represent the feature and label space of client $i$, respectively. FL is conducted by multiple rounds of global iterations, a.k.a., communication rounds. The PS is responsible for model initialization, client selection and model aggregation in each round.  Let $\theta$ denote model parameters to be learned via FL. At the beginning of communication round $t$, a subset of clients $\mathcal{S}^t$ with size $|\mathcal{S}^t| = \max(r \cdot n, 1)$ will be randomly selected to participate in FL. Here $0\le r \le1$ is the participating ratio. 
Clients in $\mathcal{S}^t$ will receive the latest model parameters $\theta^t$ from the PS, and then they perform local training for $E \in \mathbb{N}^+$ epochs. The objective of FL  is defined by  $ \min_{ \theta} \frac{1}{n} \sum_{i = 1}^{n} f_{i}(\theta)$.
Here, $f_i(\theta) = \mathbb{E}_{(x, y) \sim \mathcal{D}_i}\left[ \ell(\theta; (x, y)) \right]$ represents the empirical loss function with model $\theta$ on client $i$ and $\ell$ is the loss  defined by a single sample. After conducting local training, clients send their updated models back to the PS. The PS aggregates these local models to generate a new global model, which is then used for the next round. 
The pseudo code of the FL algorithm, \emph{i.e.}, \texttt{FedAvg} \cite{fedavg}, is presented  in Algorithm \ref{alg:fedavg}, where $SGD_{i}(\cdot)$ is 
the local optimizer to minimize $f_i(\theta)$ on the $i$-th client based on stochastic gradient descent (SGD).

Since the data distribution is non-IID in FL, implying that each $\mathcal{D}_i$ can be   drawn from a different distribution. As a result, training a uniform model $\theta$ cannot well fit heterogeneous data distributions on all clients. pFL resorts to learning personalized models denoted by $\theta_1, \theta_2, \dots, \theta_n$ for $n$ clients.

\begin{algorithm}[t]
    \caption{\texttt{FedAvg} \cite{fedavg} } \label{alg:fedavg}
    \begin{algorithmic}[0] 
        \State \textbf{Parameters:} Participating  ratio $r$, number of local epochs $E$, number of communication rounds $T$
        \State Initialized model $\theta^{0}$
        
        \State \textbf{Server Execute:}
        \State \quad \textbf{for} {$t = 1, \dots, T$} \textbf{do}
        \State \quad \quad $m \leftarrow \max(r \cdot n, 1)$; $\mathcal{S}^{t} \leftarrow $ (random set of $m$ clients)
        \State \quad \quad \textbf{for} {each client $i \in \mathcal{S}^{t}$ in parallel} \textbf{do}
        \State \quad \quad \quad $\theta_i^t \leftarrow $ ClientUpdate($i, \theta_i^{t-1}$)
        \State \quad \quad  $
            \theta^{t + 1} \leftarrow \sum_{i \in \mathcal{S}^{t}}{\frac{|\mathcal{D}_i|}{|\mathcal{D}|} \theta_{i}^{t}}, \qquad \text{where} \qquad |\mathcal{D}| = \sum_{i \in \mathcal{S}^t} |\mathcal{D}_i|$
        \\
        \State \textbf{ClientUpdate} ($i, \theta^t$) \textbf{:} \quad // \textit{run on client i}
        \State \quad Set $\theta_{i}^{t} \leftarrow \theta^{t}$
        \State \quad \textbf{for} {each local epoch} \textbf{do}
        \State \quad \quad $\theta_{i}^{t} \leftarrow SGD_{i}(\theta_{i}^{t}; \mathcal{D}_{i})$
        \State \quad \Return $\theta_{i}^{t}$
    \end{algorithmic}
\end{algorithm}

\noindent{\bf pFL by  Model Decoupling} 
Model decoupling is one of the most advanced techniques to achieve pFL by decoupling a complex neural network (NN) model as a feature extractor and a classifier. Formally, the model $\theta$ is decoupled as $\theta = \omega \circ \varphi$, where $\omega$ is the feature extractor and $\varphi$ is the classifier. 

According to previous study \cite{fedper, fedrep}, the final fully connected layer in CNN (convolutional neural network) models such as \cite{lenet5, mobilenetv2} should be included in the classifier part $\varphi$ that can well capture local data patterns. The generalization capability of CNNs is captured by other layers which should be included in the feature extractor part $\omega$. 

The objective of pFL via model decoupling is more flexible, which can be formally expressed by $ \min_{\omega} \frac{1}{n} \sum_{i=1}^{n} \min_{\varphi_{1}, \dots, \varphi_{n}} f_{i}(\omega, \varphi_{i}).$ Here $\theta_i = \omega \circ \varphi_{i}$. The inner min optimization is conducted on individual clients by tuning classifiers. Whereas, the outer min optimization is dependent on the PS by aggregating multiple feature extractors. Note that the $\omega$ part is identical for all clients and personalization is only reflected by the classifier part $\varphi_i$.

%% file: method.tex
In this section, we  conduct a measurement study to show the effectiveness to measure client similarity by the distance of classifiers. Then, we elaborate the design of the \texttt{pFedSim} algorithm. 

\subsection{Measurement Study}
In FL,  the challenge in identifying similar clients lies in the fact that data is privately owned by FL clients, which is unavailable to the PS. Through a measurement study, we unveil that the distance between classifiers, \emph{i.e.}, $\varphi_i$,  is the most effective metric to estimate client similarity. 

Our measurement study consists of three steps: 1) Classifiers can be used to precisely discriminate models trained by non-IID data.  2) The similarity between classifiers is highly correlated with the similarity of data distributions. 3) The classifier-based similarity is more effective than other metrics used in existing works.

\begin{figure}[htb]
    \centering
    \includegraphics[width=.9\textwidth]{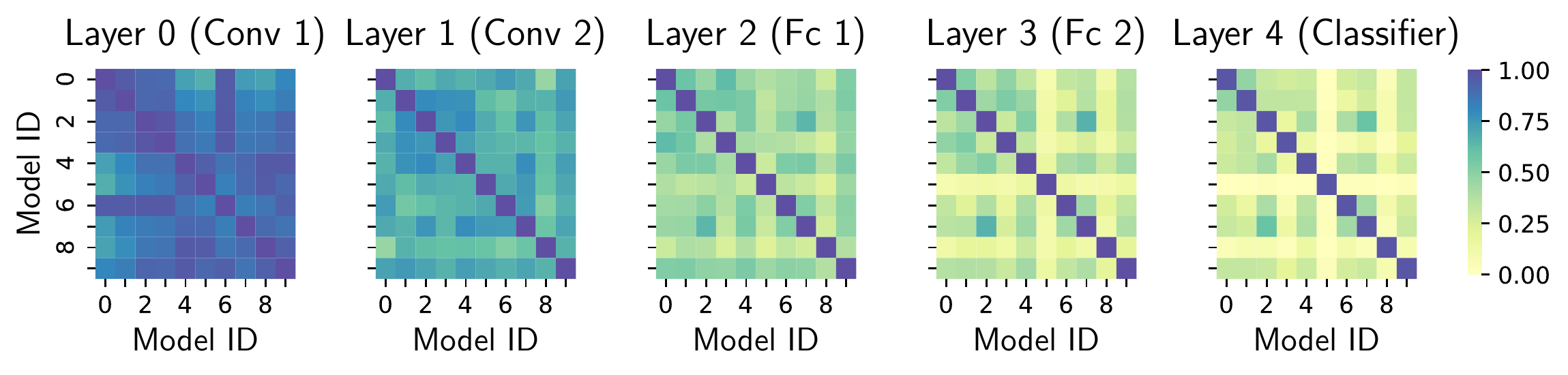}
    \caption{The comparison of CKA similarities of different layers when training 10 LeNet5 models.}
    \label{fig:cka}
    \vspace{-3mm}
\end{figure}

\noindent{\bf Step 1.} we conduct a toy experiment by using the  CIFAR-10 dataset \cite{cifar10} with labels 0 to 9, which is randomly partitioned into 10 subsets according to the Dirichlet distribution (a classical distribution to generate non-IID data) with $\alpha$ = 0.1 \cite{fedavgm}. Each subset is used to train an independent LeNet5 \cite{lenet5} model (with five layers in total) for 20 global iterations. Centered-Kernel-Alignment (CKA) proposed by \cite{cka} is used as the metric to evaluate the similarity of outputs of the same layer from two models, given the same input data. The score of CKA similarity is from 0 (totally different) to 1 (identical). We use the  CKA metric to compute the similarity for each layer between any pair of models  in Figure \ref{fig:cka}. It is interesting to note that CKA comparison of the last layer, \emph{i.e.}, the classifier, can precisely discriminate models trained on different clients.  For the comparison of Layer 0, it is almost impossible to discriminate different models. The comparison of  Layer 3 is better for discriminating models, but still worse than that of Layer 4. 

\noindent{\bf Step 2.} Although Figure \ref{fig:cka} indicates that models can be effectively distinguished through classifiers, it is still opaque how the classifier can guide us to identify similar clients.  Thus, we further conduct a toy experiment to evaluate the correlation between  classifier similarity and data distribution similarity. We define the following distance between two datasets to measure data similarity between two clients. 
\begin{equation}
    dist(\mathcal{D}_i, \mathcal{D}_j) = 1 - \frac{\left\lvert \left\{ (x, y) | y \in \mathcal{Y}_i \cap \mathcal{Y}_j \right\}_{(x, y) \sim \mathcal{D}_i \cup \mathcal{D}_j} \right\rvert}{\left\lvert\mathcal{D}_i \cup \mathcal{D}_j \right\rvert}.
    \label{def:dist}
\end{equation}
Intuitively speaking, $dist(\mathcal{D}_i, \mathcal{D}_j)$ measures the fraction of data samples belonging to labels commonly owned by clients $i$ and  $j$.  

To  visualize the distance between different clients, we employ a label non-IID setting  by manually distributing data samples of  CIFAR-10  according to their labels to 4 exclusive subsets denoted by  $\left\{ \mathcal{D}_{i}, \mathcal{D}_{i'}, \mathcal{D}_{j}, \mathcal{D}_{k}\right\}$.
$\mathcal{D}_{i}$ and $\mathcal{D}_{i'}$ contain  data samples with labels $\left\{ 0 \sim 4 \right\}$. $\mathcal{D}_{j}$ and $\mathcal{D}_{k}$ contain data samples with labels  $\left\{ 2 \sim 6 \right\}$ and $\left\{ 5 \sim 9 \right\}$, respectively. Based on Eq.~\eqref{def:dist}, it is easy to verify that:
\[
    dist(\mathcal{D}_i, \mathcal{D}_{i'}) < dist(\mathcal{D}_i, \mathcal{D}_j) < dist(\mathcal{D}_j, \mathcal{D}_k) < dist(\mathcal{D}_i, \mathcal{D}_k).
\]
We define the data similarity as $1-dist(\mathcal{D}_i, \mathcal{D}_{i'})$, and thus $sim(\mathcal{D}_i, \mathcal{D}_{i'}) > sim(\mathcal{D}_i, \mathcal{D}_j) > sim(\mathcal{D}_j, \mathcal{D}_k) > sim(\mathcal{D}_i, \mathcal{D}_k).$ For a particular dataset, \emph{e.g.}, $\mathcal{D}_i$, a model $\theta_i$ is trained independently. By decoupling $\theta_i$, let $\varphi_i$ denote its classifier. Each classifier can be further decomposed into a collection of decision boundaries, denoted by $\varphi_i = \left\{ \varphi_{i, c} \right\}_{c \in \mathcal{Y}}$, where $\varphi_{i, c}$ represents the $c$-th class decision boundary in the $i$-th client's classifier. We define the similarity between two classifiers as the average of similarities between their decision boundaries as follows:
\begin{equation}
    sim(\varphi_i, \varphi_j) = \frac{1}{\left\lvert \mathcal{Y} \right\rvert } \sum_{c \in \mathcal{Y}}{sim_{vec}(\varphi_{i, c}, \varphi_{j, c})},
\end{equation}
where $sim_{vec}(\cdot)$ is a similarity metric (\emph{e.g.}, cosine similarity in our experiment) for measuring the distance between two vectors.
We randomly initialize four LeNet5 models $\left\{\theta_{i}, \theta_{i'}, \theta_{j}, \theta_{k} \right\}$ under the same random seed and distribute them to corresponding datasets. 
Then, SGD is conducted to independently train those LeNet5 models for 20 iterations. Because of non-IID data among 4 subsets, two models will diverge after 20 iterations if their data distance is large.

\begin{figure}[t]
    \centering
    \includegraphics[width =\textwidth]{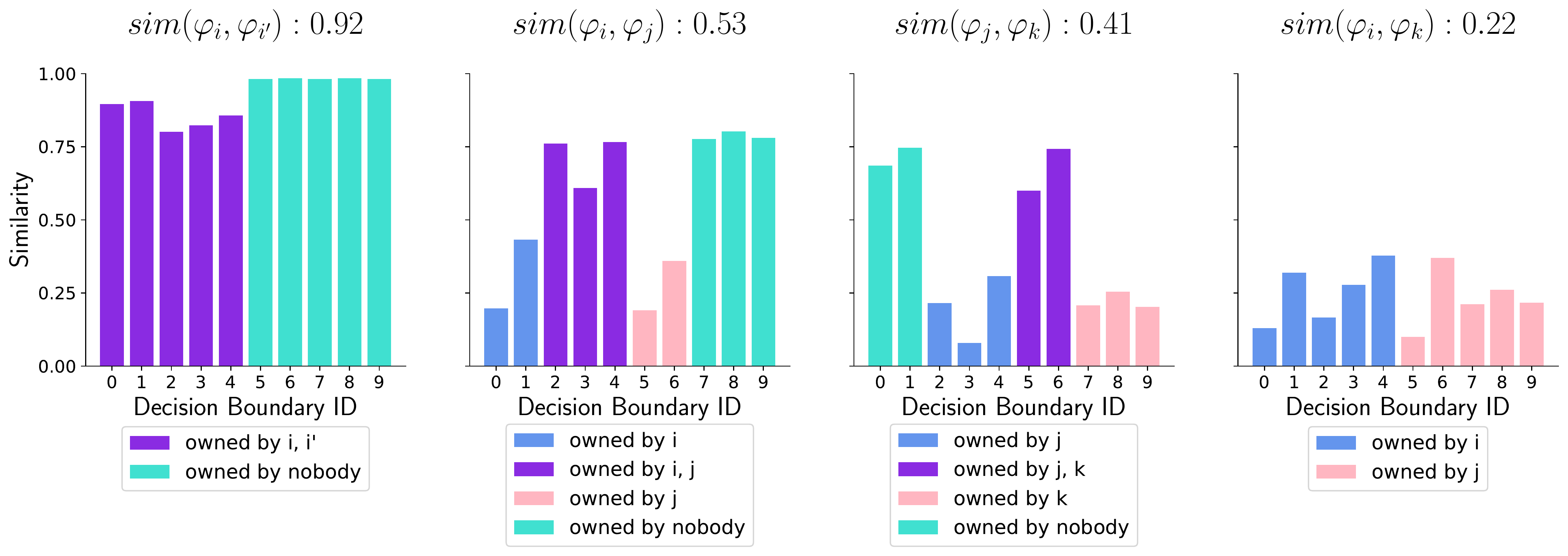}
    \caption{The cosine similarities of classifiers trained through different CIFAR-10 subsets.}
    \label{fig:cs}
    \vspace{-3mm}
\end{figure}

The experiment results are shown in Figure \ref{fig:cs}, which show similarity scores between any two classifiers together with  similarity scores between decision boundaries for each class. From the experiment results, we can observe that 
\[
    sim(\varphi_i, \varphi_{i'})>    sim(\varphi_i, \varphi_j) > sim(\varphi_j, \varphi_k) > sim(\varphi_i, \varphi_k).
\]
Recall that $sim(\mathcal{D}_i, \mathcal{D}_{i'}) > sim(\mathcal{D}_i, \mathcal{D}_j) > sim(\mathcal{D}_j, \mathcal{D}_k) > sim(\mathcal{D}_i, \mathcal{D}_k).$ This result indicates that the classifier similarity is strongly correlated with the data similarity.

We zoom in to investigate the similarity of two classifier decision boundaries for each class. 
From Figure \ref{fig:cs}, we can see that the decision boundary similarity is high only if the class label is commonly owned (or missed) from two datasets. Otherwise, the similarity score is very low if the class is only owned by a single dataset. The comparison of decision boundary similarities further verifies the effectiveness to estimate data similarity by using classifier similarity because the label distribution can heavily affect the decision boundary similarity, and hence the classifier similarity.

\setlength{\intextsep}{1mm}
\begin{wraptable}[7]{r}{.5\textwidth}
    \centering
    \resizebox{.5\textwidth}{!}{
        \begin{tabular}{l|c|c|c|c}
            Metric               & $(i, i')$            & $(i, j)$              & $(j, k)$              & $(i, k)$              \\ 
            \midrule
            $Data\ Similarity$   & 1                    & 0.53                  & 0.32                  & 0                     \\
            $WDB$ \cite{fedap}   & $1.1 \times 10^{-3}$ & $9 \times 10^{-4}$    & $1.5 \times 10^{-3}$  & $1.3 \times 10^{-3}$  \\
            $MDB$ \cite{cfl}     & 0.24                 & 0.11                  & 0.13                  & 0.08                  \\
            $LDB$ \cite{fedfomo} & $8.6 \times 10^{-6}$ & $-3.7 \times 10^{-4}$ & $-3.3 \times 10^{-4}$ & $-4.5 \times 10^{-4}$ \\
            $CS$ (Ours)          & 0.92                 & 0.53                  & 0.41                  & 0.22                  \\
        \end{tabular}
    }
    \caption{Comparison of different client similarity metrics.}
    \label{tb:metrics}
\end{wraptable}

\noindent{\bf Step 3.} To verify that classifier similarity is the most effective metric for estimating data similarity, we conduct the same experiment as Figure~\ref{fig:cs} by using metrics proposed in related works. More specifically, we implement Wasserstein distance based similarity ($WDB$) \cite{fedap}, model difference based similarity ($MDB$) \cite{cfl} and evaluation loss based similarity ($LDB$) \cite{fedfomo} based on models we have obtained in Figure~\ref{fig:cs}. $WDB$ is based on distance between outputs of batch norm layers from different models, $LDB$ is based on the difference between empirical loss evaluated on a dataset, and $MDB$ calculates the difference between models before and after model training. 
We compare data similarity, $WDB$, $MDB$, $LDB$ and our classifier similarity ($CS$) in Table \ref{tb:metrics}, from which we can observe that $CS$ is the best one to estimate data similarity because its values are highly correlated with data similarity.

Through above three steps, we have demonstrated the effectiveness to estimate data similarity by using classifier similarity. Besides, classifiers are part of model parameters exposed by clients, implying that no additional information is required. Based on our measurement findings, we design \texttt{pFedSim} since the next subsection.

\subsection{\texttt{pFedSim} Design}

Inspired by our measurement study, we propose the \texttt{pFedSim} algorithm that performs personalized model aggregation based on  the  similarity between classifiers. 
Because the initial global model is randomly generated by the PS (without factoring in local datasets), it is difficult to estimate data similarity based on initialized model parameters. Thereby,  we design   \texttt{pFedSim} with two phases:
\begin{enumerate}
    \item \textbf{Generalization Phase:} It is also called the warm-up phase. In this phase, traditional FL such as \texttt{FedAvg} is conducted to obtain a global model with a relatively effective  feature extractor and a classifier.
    \item \textbf{Personalization Phase} In this phase, we personalize feature extractors by adjusting aggregation weights based on classifier similarity. Meanwhile, classifiers are only updated by local training to better fit local data.
\end{enumerate}

We summarize the workflow of \texttt{pFedSim} and present an illustration in Appendix \ref{apx:pfedsim_workflow} to facilitate understanding, which consisting of 4 steps:
\begin{enumerate}
    \item The PS distributes the latest models to clients.
    \item Participating clients train models on their respective private datasets.
    \item Participating clients send their updated models back to the PS.
    \item In the generalization phase, the PS aggregates all received models to produce a new global model. In the personalization phase, the PS obtains the classifier similarity matrix based on uploaded model parameters. Only feature extractors similar to each other are aggregated. Then, the PS go back to Step 1 to start a new communication round.
\end{enumerate}

Note that the PS produces a single global model in the generalization phase. However, in the personalization phase, the PS produces a personalized model for each client based on the similarity matrix denoted by $\varPhi$. The final output of \texttt{pFedSim} are personalized client models $\left\{ \theta_0, \dots, \theta_n \right\}$.

\begin{algorithm}[H]
    \caption{\texttt{pFedSim}} \label{alg:pfedsim}
    \begin{algorithmic}[0] 
        \State \textbf{Parameters:} Join ratio $r$, number of local epoch $E$, number of communication round $T$, generalization ratio $\rho$, similarity matrix $\varPhi \in \mathbf{S}^{n \times n}$
        \State Initialize $\theta^0, \varPhi$
        \State Compute $T_g \leftarrow \lfloor \rho T \rfloor; T_p \leftarrow T - T_g $
        \\
        \State \textbf{Generalization Phase}
        \State Perform regular FL  \emph{e.g.}, Algorithm \ref{alg:fedavg} in $T_g$ communication rounds to obtain $\theta^{T_g - 1}$
        \\
        \State \textbf{Personalization Phase}
        \State \textbf{Server executes:}
        \State \quad Decouples $\omega^{T_g - 1}$ and $\varphi^{T_g - 1}$ from $\theta^{T_g - 1}$
        \State \quad \textbf{for} {all client $i \in \mathcal{C}$ in parallel} \textbf{do}
        \State \quad \quad Set $\omega_i^{T_g - 1} \leftarrow \omega^{T_g - 1}$
        \State \quad \quad Set $\varphi_i^{T_g - 1} \leftarrow \varphi^{T_g - 1}$
        \State \quad \textbf{for} { $t = T_g, \dots, T$ } \textbf{do}
        \State \quad \quad $m \leftarrow \max(r \cdot n, 1); \mathcal{S}^{t} \leftarrow $ (random set of $m$ clients)
        \State \quad \quad \textbf{for} {each client $i \in \mathcal{S}^{t}$ in parallel} \textbf{do}
        \State \quad \quad \quad Generate $\theta_i^t$ by Eq.~\eqref{eq:model_update}
        \State \quad \quad \quad $(\omega_i^t, \varphi_i^t) \leftarrow $ ClientUpdate($i, \theta_i^t$) 
        \State \quad \quad \textbf{for} {other client $i \notin \mathcal{S}^{t}$ in parallel} \textbf{do}
        \State \quad \quad \quad Set $\varphi_{i}^{t} \leftarrow \varphi_{i}^{t-1}$
        \State \quad \quad \textbf{for} {two different clients $i, j \in \mathcal{S}^{t}$} \textbf{do}
        \State \quad \quad \quad Compute the similarity $\varPhi_{ij}$ by Eq.~\eqref{eq:sim_calc}
        \\
        \State \textbf{ClientUpdate} ($i, \theta_i^t$) \textbf{:} \quad // \textit{run on client i}
        \State \quad Receive $\theta_i^t$ and splits it into $\omega_i^t$ and $\varphi_i^t$
        \State \quad \textbf{for} {each local epoch} \textbf{do}
        \State \quad \quad $(\omega^{t}_{i}, \varphi_{i}^{t}) \leftarrow SGD_{i}(\omega_i^{t}, \varphi_{i}^{t}; \mathcal{D}_{i})$
        \State \quad \Return ($\omega_{i}^{t}, \varphi_{i}^{t}$)
        
    \end{algorithmic}
\end{algorithm}

\noindent{\bf How to Compute Similarity}
We compute the similarity between classifiers by modifying the cosine similarity\footnote{We have tried a few different similarity metrics, and cosine similarity is the best one.}, which is symmetric, \emph{i.e.}, $sim(\varphi_i, \varphi_j) = sim(\varphi_j, \varphi_i)$, with a low computation cost. The similarity of  two classifiers is computed by:
\begin{equation}
    \varPhi_{ij} = -\frac{1}{| \mathcal{Y} |}\sum_{c \in \mathcal{Y}} \log \left[ 1  - \max \left( 0, {\frac{\varphi_{i, c} \cdot  \varphi_{j, c}}{\|\varphi_{i,c}\| \|\varphi_{j,c}\| + \epsilon }} \right) \right],
    \label{eq:sim_calc}
\end{equation}
where $\epsilon$ is always set as a small positive number (\emph{e.g.}, $10^{-8}$ in our experiments) to  avoid yielding extreme values. In our computation, the cosine similarity is further adjusted by two operations: 1) If the cosine similarity of two classifiers is negative, it makes no sense for these two clients to collaborate with each other. Thus, the final similarity is lower bounded by $0$. 2) We utilize the negative logarithm function to further adjust the cosine similarity so that clients can be easily discriminated. This operation is similar to the softmax operation in the model output layer in CNNs \cite{lenet5, mobilenetv2}.

\noindent{\bf Similarity-Based Feature Extractor Aggregation}
In the personalization phase, at the beginning of each communication round, the PS will aggregate an exclusive feature extractor for the $i$-th client in $\mathcal{S}^t$ according to the similarity matrix $\varPhi \in \mathbf{S}^{n \times n}$. The initial value of $\varPhi$ is an identity matrix. A larger value of an entry, \emph{e.g.}, $\varPhi_{ij} \in [0, 1]$, implies that  clients $i$ and $j$ are more similar to each other. The update of the personalized model for the $i$-th client in the personalization phase is:
\begin{equation}
    \theta_i^t = \varphi_i^t \circ \omega_i^t =
    \begin{cases}
        \omega_i^t  =  \frac{1}{\sum_{j\in \mathcal{C}}{\varPhi_{ij}}} \sum_{j\in \mathcal{C}}{\varPhi_{ij} \omega_j^{t-1}}, &  \\
        \varphi_i^t  =  \varphi_i^{t-1}.                                                                                     & 
    \end{cases}
    \label{eq:model_update}
\end{equation}
Note that personalization is guaranteed because: 1) $\varphi_i$ is only updated locally without aggregating with others. 2) $\varPhi_{ij}$ (computed based on $\varphi_i$ and $\varphi_j$) is incorporated for personalizing the aggregation of feature extractors.

We summarize \texttt{pFedSim} in Algorithm \ref{alg:pfedsim}. Our method \texttt{pFedSim} does not depend on exchanging of additional information rather than model parameters. 
Therefore, the advantage of \texttt{pFedSim} includes a lower cost and a higher privacy protection level, compared with other baselines such as \cite{fedap, fedfomo}.

%% file: experiment.tex
In this section, we report our experiment results conducted with real datasets to demonstrate the superb performance of \texttt{pFedSim}. 

\subsection{Experiment Setup}

\begin{wrapfigure}[16]{r}{.3\textwidth}
    \centering
    \includegraphics[width=\linewidth]{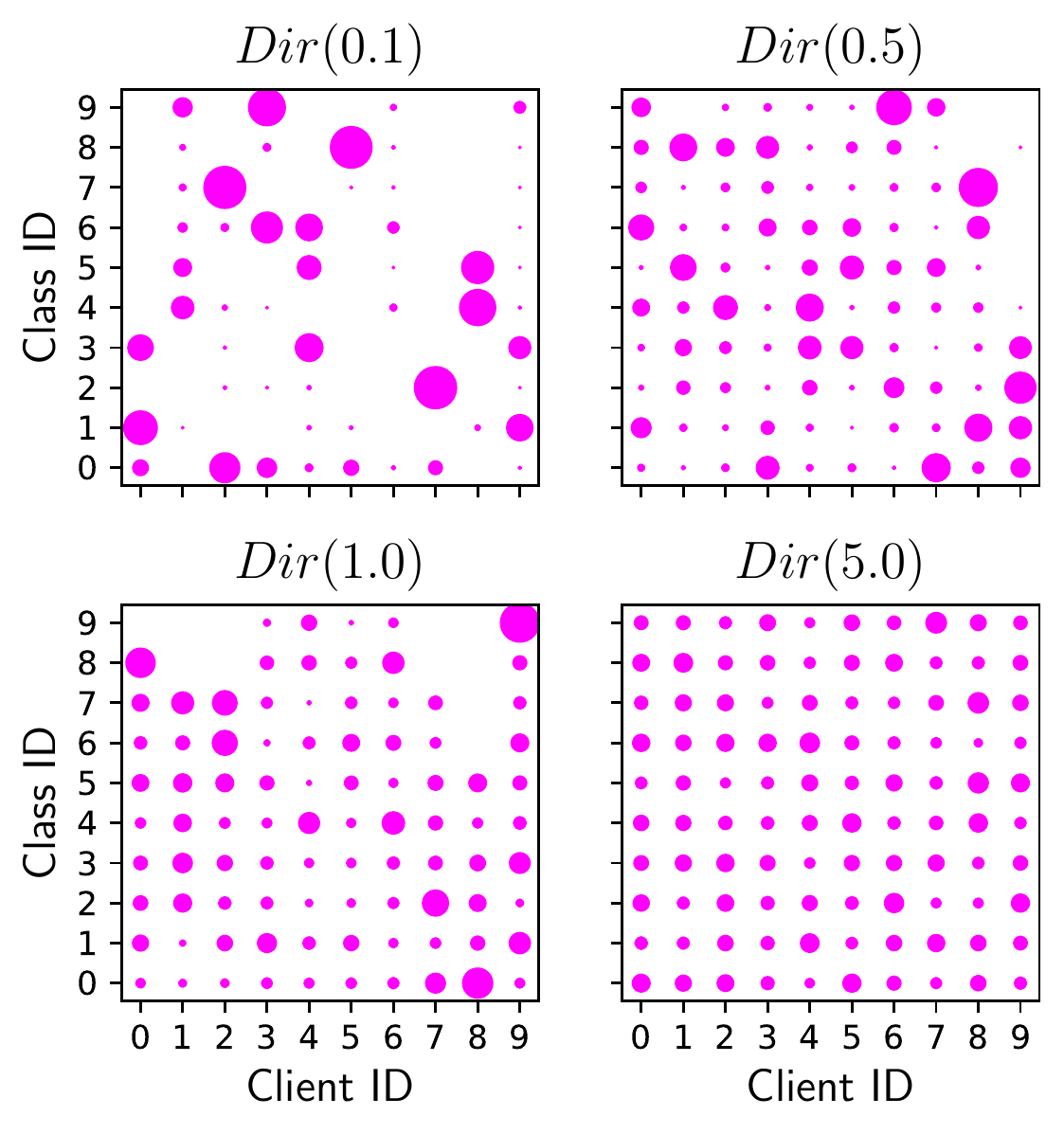}
    \caption{Data distribution of CIFAR-10 images allocated to clients with different $\alpha$.}
    \label{fig:dir}
\end{wrapfigure}

\noindent{\bf Datasets}\label{sec:dataset}
We evaluate  \texttt{pFedSim} on four standard image datasets, which are CIFAR-10 \cite{cifar10}, CINIC-10 \cite{cinic10}, EMNIST \cite{emnist} and Tiny-ImageNet \cite{tinyimgnet}. We split each dataset into 100 subsets (\emph{i.e.}, $n = 100$) according to the Dirichlet distribution $Dir(\alpha)$ with $\alpha \in \left\{ 0.1, 0.5 \right\}$ to  simulate  scenarios with non-IID data \cite{fedavgm}. The non-IID degree  is determined  by $\alpha$. When $\alpha$ is very small, \emph{e.g.}, $\alpha = 0.1$, the data non-IID degree is more significant implying  that  the data owned by a particular client cannot cover all  classes, \emph{i.e.}, $|\mathcal{Y}_i| \le |\mathcal{Y}|$, where $\mathcal{Y}_i$ is the label space of data distributed on the $i$-th client. Figure \ref{fig:dir} visualizes an example of the data distribution partitioned according to  the $Dir(\alpha)$ distribution. The x-axis represents the client ID while the y-axis represents the class ID. The circle size represents the number of samples of a particular class allocated to a client. As we can see, the non-IID degree is more significant when $\alpha$ is smaller. More dataset details are presented in Appendix \ref{apx:dataset}.

\setlength{\intextsep}{-1mm}

\noindent{\bf Baselines}\label{sec:baseline}
We compare \texttt{pFedSim} with both FL and state-of-the-art pFL baselines. In order to create a baseline for evaluating the generalization and personalization performance, respectively, we implement \texttt{FedAvg} \cite{fedavg} and Local-training-only in our experiments.
Additionally, we implement the following baselines for comparison: 
1) \texttt{FedProx} \cite{fedprox} is a regular FL method that adds a proximal term to the loss function as a regularization method to prevent the local model from drifting away from the global model; 
2) \texttt{FedDyn} \cite{feddyn} is a regular FL method that utilizes a dynamic regularizer for each client to align the client local model with the global model; 
3) \texttt{FedGen} \cite{fedgen} is a regular FL method that trains a feature generator to generate virtual features for improving local training; 
4) \texttt{FedPer} \cite{fedper} is a pFL method that preserves the classifier of each client model locally to achieve personalization; 
5) \texttt{FedRep} \cite{fedrep} is a pFL method that preserves the classifier locally, and trains the classifier and the feature extractor sequentially; 
6) \texttt{FedBN} \cite{fedbn} is a pFL method that preserves batch normalization layers in the model locally to stabilize training; 
7) \texttt{Per-FedAvg} \cite{perfedavg} is a pFL method that combines first-order meta-learning for quick model adaptation after local fine-tuning; 
8) \texttt{FedFomo} \cite{fedfomo} is a pFL method that personalizes model aggregation based on the loss difference between client models; 
9) \texttt{FedAP} \cite{fedap} is a pFL method using the Wasserstein distance as the metric to measure the client-wised similarity, which is then used to  optimize model aggregation.

\noindent{\bf Implementation}\label{exp_impl}
We implemented LeNet5 \cite{lenet5}  for all methods to conduct performance evaluation on CIFAR-10, CINIC-10 and EMNIST. MobileNetV2 \cite{mobilenetv2} is implemented to classify Tiny-ImageNet images. According to the setup in \cite{fedap}, we evenly split the dataset of  each client into the training and test sets with no intersection, \emph{i.e.}, $|\mathcal{D}_{i}^{train}| = |\mathcal{D}_{i}^{test}|$. On the client side, we adopt SGD as the local optimizer. The local learning rate is set to 0.01 for all experiments like \cite{fedap}. All experiments shared the same join ratio $r = 0.1$, communication round $T = 200$, local epoch $E = 5$, and batch size 32. For \texttt{pFedSim}, we set the generalization ratio $\rho = 0.5$ (\emph{i.e.}, $T_g = 100, T_p = 100$). 
We list the used model architectures and full hyperparameter settings among aforementioned methods in Appendix \ref{apx:model}, \ref{apx:hyper} respectively.

\begin{table}[t!]
    \setlength{\abovecaptionskip}{3mm}
    \centering
    \resizebox{\textwidth}{!}{
        \begin{tabular}{lcc|cc|cc|cc}
            \toprule
            \multicolumn{1}{c}{ }                & \multicolumn{2}{c}{CIFAR-10} & \multicolumn{2}{c}{CINIC-10} & \multicolumn{2}{c}{Tiny-ImageNet} & \multicolumn{2}{c}{EMNIST}                                                                                                             \\
            \cmidrule(lr){2-3}\cmidrule(lr){4-5} \cmidrule(lr){6-7} \cmidrule{8-9}
            \#Partition                          & $Dir(0.1)$                   & $Dir(0.5)$                   & $Dir(0.1)$                        & $Dir(0.5)$                 & $Dir(0.1)$               & $Dir(0.5)$               & $Dir(0.1)$               & $Dir(0.5)$               \\
            \midrule
            Local-Only                           & {84.30 (0.81)}               & {56.80 (0.80)}               & {83.13 (1.68)}                    & {54.51 (1.11)}             & {43.70 (0.44)}           & {19.65 (0.08)}           & {93.54 (0.43)}           & {84.17 (0.06)}           \\
            \texttt{FedAvg} \cite{fedavg}        & {24.77 (4.25)}               & {40.90 (2.48)}               & {27.28 (1.83)}                    & {39.44 (0.95)}             & {27.66 (6.35)}           & {41.45 (0.13)}           & {75.27 (1.55)}           & {82.17 (0.27)}           \\
            \texttt{FedProx} \cite{fedprox}      & {29.70 (2.29)}               & {41.81 (2.46)}               & {26.23 (3.83)}                    & {40.63 (2.09)}             & {29.47 (0.74)}           & {40.98 (0.34)}           & {74.22 (1.43)}           & {81.81 (0.15)}           \\
            \texttt{FedDyn} \cite{feddyn}        & {34.97 (0.75)}               & {44.27 (1.25)}               & {25.44 (1.50)}                    & {37.50 (1.75)}             & {25.09 (0.78)}           & {27.04 (1.27)}           & {72.39 (1.81)}           & {80.04 (0.68)}           \\
            \texttt{FedGen} \cite{fedgen}        & {23.51 (4.27)}               & {43.51 (1.28)}               & {21.18 (1.25)}                    & {39.67 (1.43)}             & {31.04 (0.48)}           & \underline{42.43 (0.40)} & {75.97 (0.85)}           & {83.20 (0.23)}           \\
            \texttt{FedBN} \cite{fedbn}          & {40.75 (5.82)}               & {47.44 (2.17)}               & {43.27 (2.03)}                    & {44.51 (1.15)}             & {29.51 (4.04)}           & {40.00 (1.25)}           & {77.45 (1.40)}           & {82.28 (0.60)}           \\
            \texttt{Per-FedAvg} \cite{perfedavg} & {75.26 (3.37)}               & {44.96 (1.99)}               & {78.29 (2.95)}                    & {52.66 (0.44)}             & {34.04 (1.04)}           & {23.57 (1.75)}           & {93.04 (0.96)}           & {86.90 (0.09)}           \\
            
            \texttt{FedPer} \cite{fedper}        & {82.42 (0.55)}               & {62.37 (0.90)}               & {81.94 (1.63)}                    & {61.08 (1.10)}             & {52.13 (0.80)}           & {27.27 (0.30)}           & {94.42 (0.47)}           & {87.50 (0.22)}           \\
            
            \texttt{FedRep} \cite{fedrep}        & \underline{84.85 (0.74)}     & {64.47 (1.03)}               & \underline{84.30 (1.34)}          & \underline{64.42 (1.35)}   & \underline{57.28 (0.63)} & {33.57 (0.40)}           & \underline{95.12 (0.44)} & {88.38 (0.24)}           \\
            
            \texttt{FedFomo} \cite{fedfomo}      & {84.70 (0.72)}               & {58.14 (0.96)}               & {82.42 (1.93)}                    & {54.61 (1.45)}             & {40.80 (0.48)}           & {23.72 (1.76)}           & {93.20 (0.46)}           & {83.43 (0.06)}           \\
            
            \texttt{f-FedAP} \cite{fedap}        & {84.13 (0.76)}               & \underline{64.73 (0.62)}     & {82.84 (1.58)}                    & {62.21 (0.85)}             & {42.70 (0.25)}           & {41.69 (1.49)}           & {94.34 (0.40)}           & \underline{89.00 (0.06)} \\
            \midrule
            \texttt{pFedSim} (Ours)              & \textbf{86.76 (0.84)}        & \textbf{67.34 (0.36)}        & \textbf{84.34 (1.25)}             & \textbf{64.42 (0.77)}      & \textbf{64.91 (0.60)}    & \textbf{52.23 (0.16)}    & \textbf{95.70 (0.31)}    & \textbf{90.18 (0.11)}    \\
            \bottomrule 
        \end{tabular}
    }
    \caption{Average model accuracy over datasets with format mean(std). Bold, underline mean the best, second-best results respectively.}
    \label{tb:res}
    \vspace{-9mm}
\end{table}

\subsection{Experiment Results}
We evaluate \texttt{pFedSim} from three perspectives: model accuracy, effect of the hyperparameter $\rho$ and overhead. The experiment results demonstrate that: 1) \texttt{pFedSim} achieves the highest model accuracy in non-IID scenarios; 2) The accuracy performance of \texttt{pFedSim} is not sensitive to the hyperparameter. 3)  Low overhead is the advantage of \texttt{pFedSim}. 
The extra experiments such as the effect of hyperparameters and overhead evaluation are presented in Appendix \ref{apx:extra_exp}.

\noindent{\bf Comparing Classification Accuracy} We compare the average model accuracy performance of \texttt{pFedSim} with other baselines in Table \ref{tb:res}. Based on the experiment results, we can observe that: 
1) Our proposed algorithm, \texttt{pFedSim}, significantly outperforms other methods  achieving the highest model accuracy in all cases. 
2) When $\alpha=0.1$ indicating a significant non-IID degree of data distribution, the performance of pFL algorithms is better because of their capability to handle non-IID data. In particular, the improvement of \texttt{pFedSim} is more significant when $\alpha=0.1$.  In contrast, FL methods such as \texttt{FedAvg} deteriorate considerably when $\alpha$ is small. 
3) The \texttt{f-FedAP} algorithm also includes  generalization and personalization phases. Its model aggregation is based on the similarity of  the output of batch norm layers extracted from clients, which may  increase the data privacy leakage risk. Moreover, its model accuracy performance is worse than ours indicating that the classifier distance is more effective for estimating client similarity. 
4) Classifying the Tiny-ImageNet dataset under our experimental setting is a challenging task because  each client only owns 550 training samples on average,  but needs to solve a 200-class classification task. In this extreme case, it is worth noting that \texttt{pFedSim} remarkably improves model accuracy performance by almost 10\% at most.

%% file: conclusion.tex
Data heterogeneity is one of the biggest challenges hampering the development of FL with high model utility. Despite that significant efforts have  been made to solve this challenge through pFL, an efficient, secured and accurate pFL algorithm is still absent.
In this paper, we conduct a measure study to show the effectiveness to identify similar clients based on the classifier-based distance. Accordingly, we propose a novel pFL algorithm called \texttt{pFedSim} that only leverages the classifier-based similarity to conduct personalized model aggregation, without exposing additional information or incurring extra communication overhead. Through extensive experiments on four real image datasets, we  demonstrate that \texttt{pFedSim} outperforms other FL methods by improving model accuracy 2.96\% on average compared with state-of-the-art baselines. Besides, the \texttt{pFedSim} algorithm is of enormous practical value because its overhead cost is low and its performance insensitive  to tuning the hyperparameter.

%% file: appendix.tex
\section{Experiment Details}

\subsection{Computing Configuration}
We implement our experiments using PyTorch \cite{pytorch}, and run all experiments on Ubuntu 18.04.5 LTS with the configuration of Intel(R) Xeon(R) Gold 6226R CPU@2.90GHz and NVIDIA GeForce RTX 3090 to build up the entire workflow.

\subsection{\texttt{pFedSim} Workflow Illustration}\label{apx:pfedsim_workflow}
\begin{figure}[H]
    \vspace{5mm}
    \setlength{\abovecaptionskip}{3mm}
    \centering
    \includegraphics[width=\linewidth]{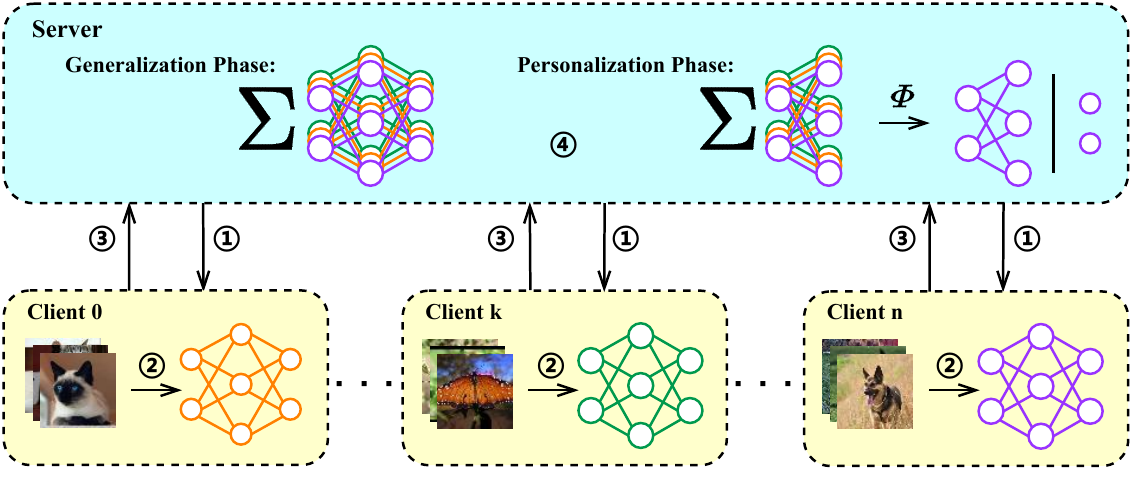}
    \caption{The execution  workflow of \texttt{pFedSim}.}
    \label{fig:flow}
\end{figure}

The \texttt{pFedSim} method is implemented by conducting the following steps for multiple rounds.  1) The PS distributes the latest models to participating clients at the beginning of each round; 2) Each client updates the received model by local training with their private dataset; 3) Each client sends their trained model parameters back to the PS. 4) In the generalization phase, the PS will aggregate all received models from clients and output a new single model as the target for the next communication round. In the personalization phase, the PS will aggregate an exclusive model for each client participating in the next communication round based on the similarity matrix $\varPhi$. The $\varPhi$ will be updated by calculating the cosine similarity of pairwise classifiers from models submitted by participating clients.

\subsection{Datasets Details}\label{apx:dataset}
\begin{table}[H]
    \vspace{3mm}
    \setlength{\abovecaptionskip}{3mm}
    \centering
    \begin{tabular}{lcccc}
        \toprule
        \multirow{2}{*}{Dataset}        & \multirow{2}{*}{Number of Samples} & \multirow{2}{*}{Number of Classes} & \multicolumn{2}{c}{Distribution Mean (std)}              \\
                                        &                                    &                                    & $Dir(0.1)$                                  & $Dir(0.5$) \\
        \cmidrule(lr){1-5}
        CIFAR-10 \cite{cifar10}         & 60,000                             & 10                                 & 600 (487)                                   & 600 (170)  \\
        CINIC-10 \cite{cinic10}         & 270,000                            & 10                                 & 2700 (1962)                                 & 2700 (843) \\
        EMNIST \cite{emnist}            & 805,263                            & 62                                 & 8142 (2545)                                 & 8142 (782) \\
        Tiny-ImageNet \cite{tinyimgnet} & 110,000                            & 200                                & 1100 (118)                                  & 1100 (38)  \\
        \bottomrule
    \end{tabular}
    \caption{Statistical information of used datasets and their distribution information on clients. }
    \vspace{1mm}
    \label{tb:dataset}
\end{table}

We list the basic information of datasets we used for our experiments in Table \ref{tb:dataset}. To make a fair comparison between different methods, we partition each dataset into 100 subsets (\emph{i.e.}, 100 clients) and the data distributions are fixed in our experiments when comparing different methods.

\subsection{Model Architecture}\label{apx:model}

We list the detailed architecture of LeNet5 \cite{lenet5} and MobileNetV2 \cite{mobilenetv2} in this section in Table \ref{tb:lenet_arch} and Table \ref{tb:mobile_arch}, respectively, which are the model backbone we used in our experiments. We list layers with abbreviations in PyTorch style. For MobileNetV2, we use the built-in API in TorchVision \cite{torchvision} with its default pretrained model weights.
\begin{table}[t]
    \setlength{\abovecaptionskip}{5mm}
    \vspace{5mm}
    \centering
    \begin{tabular}{ll}
        \toprule
        Component  & Layer                                                     \\
        \cmidrule(lr){1-2}

        \multirow{6}{*}{\shortstack{Feature                                    \\Extractor}}           & Conv2D($in$=3, $out$=6, $kernel$=5, $stride$=1, $pad$=0)  \\
                   & Conv2D($in$=3, $out$=16, $kernel$=5, $stride$=1, $pad$=0) \\
                   & MaxPool2D($kernel$=2, $stride$=2)                         \\
                   & Flatten()                                                 \\
                   & FC($out$=120)                                             \\
                   & FC($out$=84)                                              \\
        \cmidrule(lr){1-2}
        Classifier & FC($out$=$num\_classes$)                                  \\
        \bottomrule
    \end{tabular}
    \caption{LeNet5 Architecture. Conv2D() consists of a 2D convolution layer, batch normalization layer, and ReLU activation, which is executed sequentially. The $in, out, kernel, stride, pad$ represent the input channel, output channel, kernel size, convolution kernel step size moving, padding size, respectively; MaxPool2D() is a max pooling layer for 2D input; Flatten() is for reshaping input from 2D to 1D; FC() is a fully connected layer, $out$ means the number of features output has.}
    \label{tb:lenet_arch}
    \vspace{-3mm}
\end{table}

\subsection{Full Hyperparameter settings}\label{apx:hyper}
We list all hyperparameter settings among all the aforementioned methods here. Most hyperparameter settings of baselines are consistent with the values in their own papers.
\begin{itemize}
    \item \texttt{FedProx} \cite{fedprox} We set the $\mu = 1$.
    \item \texttt{FedDyn} \cite{feddyn} We set the $\alpha = 0.01$.
    \item \texttt{FedGen} \cite{fedgen} We set the hidden dimension of generator to $32$; the noise dimension to $32$; the input/output channels, latent dimension is adapted to the datasets and model backbone; the number of iterations for training the generator in each communication round to $5$; the batch size of generated virtual representation to $32$; $\alpha_{\text{generative}}$ and $\alpha_{\text{generative}}$ to $10$ and decay with factor $0.98$ in the end of each communication round.
    \item \texttt{Per-FedAvg} \cite{perfedavg} We set the $\alpha = 0.01$ and $\beta = 0.001$.
    \item \texttt{FedRep} \cite{fedrep} We set the epoch for training feature extractor part to 1.
    \item \texttt{FedFomo} \cite{fedfomo} We set the $M = 5$ and the ratio of validation set to $0.2$.
    \item  \texttt{FedAP} \cite{fedap} We implement the version called \texttt{f-FedAP}, which  performs \texttt{FedBN} \cite{fedbn} for warming up without relying on pre-training, with a generalization ratio of $0.5$ and model momentum $\mu = 0.5$.
    \item \texttt{pFedSim} (Ours) We set the generalization ratio $\rho = 0.5$.
\end{itemize}

\section{Supplementary Experiments}\label{apx:extra_exp}
We present more experimental results in the supplementary document, which are conducted for evaluating \texttt{pFedSim} more sufficiently.
\subsection{Role of Generalization Phase}\label{apx:warmup}
\begin{table}[t]
    \setlength{\abovecaptionskip}{4mm}
    \centering
    \resizebox{\textwidth}{!}{
        \begin{tabular}{lcc|cc|cc|cc}
            \toprule
            \multicolumn{1}{c}{ }        & \multicolumn{2}{c}{CIFAR-10 (LeNet5)} & \multicolumn{2}{c}{CINIC-10 (LeNet5)} & \multicolumn{2}{c}{Tiny-ImageNet  (MobileNetV2)} & \multicolumn{2}{c}{EMNIST (LeNet5)}                                                                                                 \\
            \cmidrule(lr){2-3}\cmidrule(lr){4-5} \cmidrule(lr){6-7} \cmidrule{8-9}
            \#Partition                  & $Dir(0.1)$                            & $Dir(0.5)$                            & $Dir(0.1)$                                       & $Dir(0.5)$                          & $Dir(0.1)$            & $Dir(0.5)$            & $Dir(0.1)$            & $Dir(0.5)$            \\
            \midrule
            $\rho = 0$                   & {81.61 (1.52)}                        & {56.76 (0.57)}                        & {81.70 (1.03)}                                   & {81.70 (1.03)}                      & {55.72 (0.65)}        & {32.72 (0.25)}        & {94.91 (0.53)}        & {87.92 (0.16)}        \\
            \midrule
            $\rho = 0.1$                 & {83.29 (1.64)}                        & {61.73 (0.89)}                        & {83.31 (2.12)}                                   & {60.98 (0.89)}                      & {62.30 (0.54)}        & {47.29 (0.15)}        & {95.12 (0.46)}        & {88.72 (0.15)}        \\
            $\rho = 0.3$                 & {85.89 (1.28)}                        & {65.36 (0.79)}                        & {84.15 (1.74)}                                   & {62.74 (0.82)}                      & \textbf{65.53 (0.50)} & \textbf{52.50 (0.28)} & {95.49 (0.41)}        & {89.71 (0.15)}        \\
            $\rho = 0.5$                 & {86.76 (0.84)}                        & {67.34 (0.36)}                        & {84.34 (1.25)}                                   & {64.42 (0.77)}                      & {64.91 (0.60)}        & {52.23 (0.16)}        & {95.70 (0.31)}        & {90.18 (0.11)}        \\
            $\rho = 0.7$                 & \textbf{87.05 (0.87)}                 & \textbf{68.09 (0.62)}                 & \textbf{85.13 (1.17)}                            & \textbf{64.87 (0.73)}               & {61.71 (0.65)}        & {50.63 (0.51)}        & \textbf{95.81 (0.39)} & \textbf{90.58 (0.12)} \\
            $\rho = 0.9$                 & {79.05 (2.59)}                        & {64.70 (1.43)}                        & {78.54 (1.92)}                                   & {61.83 (0.47)}                      & {52.55 (0.22)}        & {47.16 (0.34)}        & {92.89 (0.58)}        & {89.66 (0.36)}        \\
            \midrule
            $\rho = 1$ (\texttt{FedAvg}) & {24.77 (4.25)}                        & {40.90 (2.48)}                        & {27.28 (1.83)}                                   & {39.44 (0.95)}                      & {27.66 (6.35)}        & {41.45 (0.13)}        & {75.27 (1.55)}        & {82.17 (0.27)}        \\
            \bottomrule
        \end{tabular}
    }
    \caption{Average model accuracy of \texttt{pFedSim} with format mean(std) and different $\rho$.}
    \label{tb:warmup}
    \vspace{-3mm}
\end{table}
In this experiment, we vary $\rho$ from $0$ (without generalization phase) to $1$, and then compare the average model accuracy of \texttt{pFedSim}. When $\rho=1$, \texttt{pFedSim} is degenerated to \texttt{FedAvg}, and when $\rho=0$, \texttt{pFedSim} will not be warmed up. We consider those two cases as control cases to fully demonstrate the importance of the generalization phase. Note that the performance with $\rho=1$ is always the worst in all cases because \texttt{FedAvg} is incapable of handling severe data non-IID situations, while the model performance with $\rho=0$ is always worse than other cases with $\rho \in (0, 1)$ in all experiment settings because the trained model is short in generalization capability. However, an oversized generalization phase (\emph{i.e.}, $\rho=0.9$) would result in insufficient personalization and ultimately degradation of the final performance. Thus $\rho$ should be set within a proper range. Although it is not easy to determine exactly the optimal value of $\rho$, which is related to the trained model, we can draw the following observations from our experiments. For large models (\emph{e.g.}, MobileNetV2), it is better to set a smaller $\rho$, implying that the personalization of large models consumes more communication rounds. For small models (\emph{e.g.}, LeNet5), it is better to set a relatively large $\rho$ because personalization can be completed faster.
More importantly, the accuracy performance is very close to the highest performance and stable when $\rho$ is in $[0.3, 0.7]$ for all cases, making setting $\rho$ easy in practice.

\subsection{Overhead Comparison}\label{apx:overhead}
\begin{table}[H]
    \vspace{3mm}
    \setlength{\abovecaptionskip}{3mm}
    \centering
    \begin{tabular}{lcccc|c}
        \toprule
        \multicolumn{1}{c}{ }           & \multicolumn{4}{c}{$\xi_p$ $(\times 10^{-4}s)$ with format mean(std)} & \multicolumn{1}{c}{$\xi_m$}                                                                  \\
        \cmidrule(lr){2-5}\cmidrule{6-6}
        \#Dataset                       & CIFAR-10                                                              & CINIC-10                    & Tiny-ImageNet  & EMNIST         & -                            \\
        \midrule
        \texttt{FedAvg} \cite{fedavg}   & {8.68 (0.16)}                                                         & {9.59 (0.38)}               & {57.60 (1.65)} & {7.85 (0.42)}  & $\mathcal{O}(\nu_m)$         \\
        \texttt{FedFomo} \cite{fedfomo} & {11.28 (0.21)}                                                        & {11.45 (0.27)}              & {73.40 (4.15)} & {9.98 (0.37)}  & $\mathcal{O}(M \nu_m)$       \\
        \texttt{FedGen} \cite{fedgen}   & {10.56 (0.13)}                                                        & {11.18 (0.49)}              & {61.73 (2.67)} & {10.07 (0.46)} & $\mathcal{O}(\nu_m + \nu_g)$ \\
        \texttt{pFedSim} (Ours)         & {8.69 (0.17)}                                                         & {9.56 (0.48)}               & {57.64 (1.80)} & {7.89 (0.37)}  & $\mathcal{O}(\nu_m)$         \\
        \midrule
    \end{tabular}
    \caption{Overheads comparison.}
    \label{tb:overhead}
\end{table}

We compare computation and communication overhead between \texttt{pFedSim} with three most representative baselines, \emph{i.e.}, \texttt{FedAvg}, \texttt{FedFomo} and \texttt{FedGen}, in Table \ref{tb:req} to demonstrate the superiority of \texttt{pFedSim}. We split datasets according to $Dir(0.1)$ and follow the experiment setting  $T = 20$, $r = 0.1$ (\emph{i.e.}, $|\mathcal{S}|$ = 10), $E = 5$.
The computation overhead is based on the measured actual running time of each method. The communication overhead is based on the complexity of exchanging model parameters via communications, which is related to model size.
To explicitly compare communication overhead, let $\nu$ denote the number of parameters of a neural network model, and then $\nu_m$ simply represent the number of parameters of the model trained by FL. \texttt{FedGen} \cite{fedgen} is a special method requiring an additional generator model for generating virtual features to assist in  training, which has a relatively simple architecture (\emph{e.g.}, multi-layer perceptron).
Let $\nu_g$ denote the size of the generator.
Finally, we show the comparison results in Table \ref{tb:overhead}.

\noindent{\bf Computation Overhead}
We quantize the computation overhead by the average running time of the local training procedure on clients, which is denoted by
\[\xi_{p} = \sum_{t = 0}^{T} \sum_{i \in \mathcal{S}^t} \frac{\textit{LocalTrainingTime}_{i}} { T|\mathcal{S}^t| |\mathcal{D}_i^{train}|}.
\]
We record local training time in 5 repetitions and report mean and standard deviation. As we can see, $\xi_p$ of \texttt{pFedSim} is very close to \texttt{FedAvg}, indicating that \texttt{pFedSim} will not involve extra computation overhead compared to the most basic FL method; \texttt{FedFomo} \cite{fedfomo} splits a validation set from the training set additionally for evaluating the performance of models transmitted from the PS that submitted from $M$ neighboring clients. The evaluation result is then used to guide the personalized model aggregation at the client side. Therefore, it is necessary to consider both training and validation time costs for \texttt{FedFomo}. For simplicity, we simply compare the average processing time cost per sample by each method.

\noindent{\bf Communication Overhead} Let $\xi_m$ denote the communication overhead. Because real communication time is influenced by realistic network conditions, to fairly compare different methods by avoiding the randomness influence from networks, we compare the communication complexity of different methods. To run the most fundamental \texttt{FedAvg} method, each client only needs to transmit $\nu_m$ parameters implying that its communication complexity is $\mathcal{O}(\nu_m)$. Notably, the communication complexity of \texttt{pFedSim} is also $\mathcal{O}(\nu_m)$ because \texttt{pFedSim} does not need to transmit any additional information from clients to the PS.
In contrast, both \texttt{FedFomo} and \texttt{FedGen} consume heavier communication overhead. For \texttt{FedFomo}, the PS needs to transmit additional $M$ models of $M$ neighboring clients to each client, and consequently its communication overhead is extremely heavy with $\mathcal{O}(M\nu_m)$. For \texttt{FedGen}, each client needs to transmit an additional generator other than $\nu_m$ model parameters. and hence its communication overhead is $\mathcal{O}(\nu_m + \nu_g)$.

\subsection{Evaluation over DomainNet}

\begin{figure}[H]
    \centering
    \vspace{3mm}
    \begin{subfigure}[b]{.55\textwidth}
        \includegraphics[width=\textwidth]{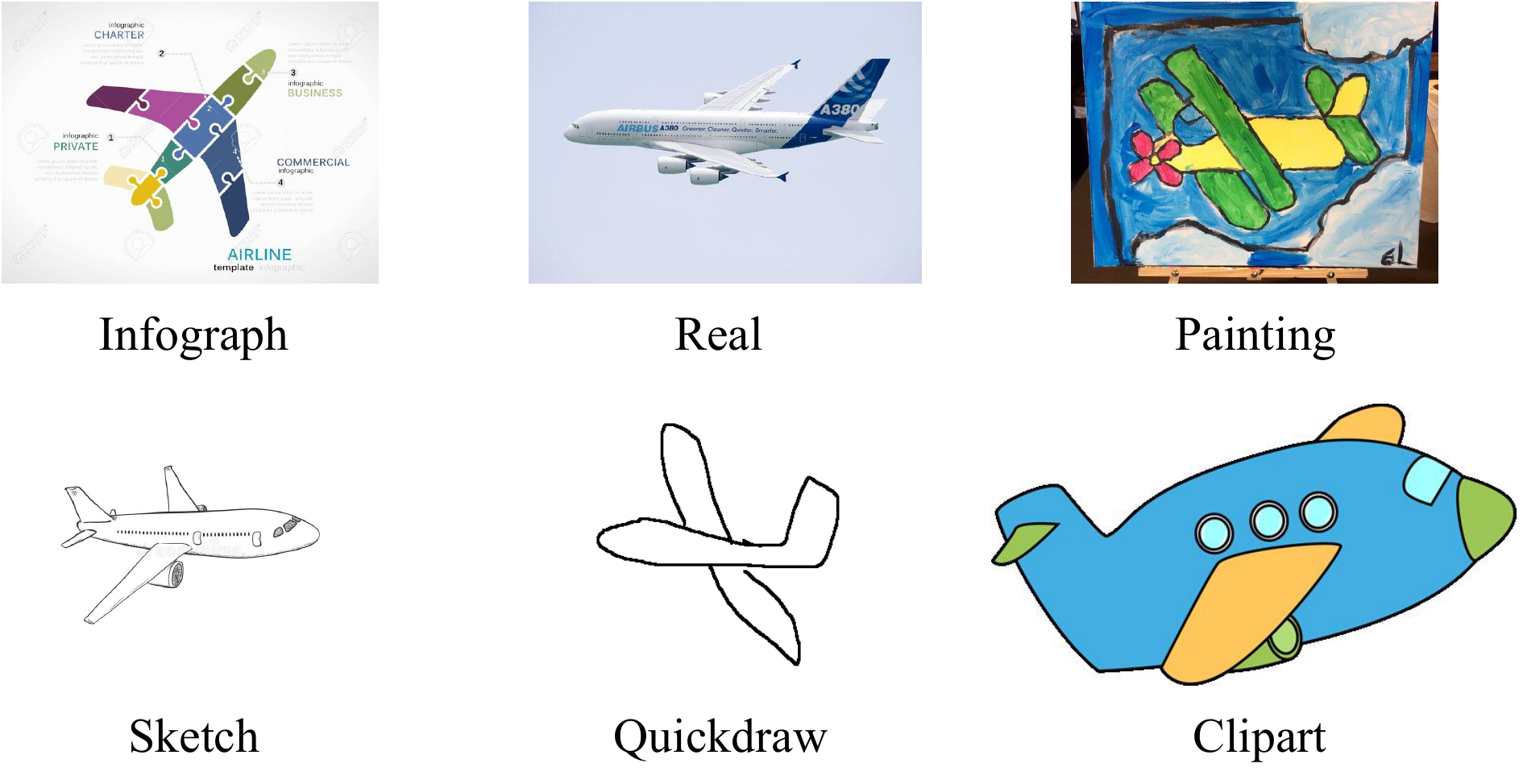}
        \subcaption{Example to show the feature non-IID with Airplane images of DomainNet  from different domains.}
        \label{fig:domain_imgs}
    \end{subfigure}
    \quad
    \begin{subtable}[b]{.4\textwidth}
        \begin{tabular}{l|c}
            Domains   & Number of Samples \\
            \midrule
            Clipart   & 2464              \\
            Infograph & 3414              \\
            Painting  & 5006              \\
            Quickdraw & 10000             \\
            Real      & 10609             \\
            Sketch    & 5056              \\
        \end{tabular}
        \subcaption{Number of data sampled from each domain in DomainNet.}
        \label{tb:domain}
    \end{subtable}
\end{figure}

We just evaluate all these methods under the label non-IID setting until now. For comprehensively demonstrating the superiority of \texttt{pFedSim}, we additionally train and evaluate aforementioned methods over DomainNet \cite{domainnet}, in which the non-IID nature of data can be divided into two categories: label non-IID and feature non-IID. The former indicates that each FL client's private data does not cover all labels, which is the basis for the data partition method used in Section \ref{sec:dataset}. The latter is introduced by \cite{fedbn} and is the intrinsic nature of DomainNet. In Figure \ref{fig:domain_imgs}, we show image samples from DomainNet that belong to the same class to vividly illustrate the feature non-IID nature.

\begin{table}[H]
    \centering
    \setlength{\abovecaptionskip}{3mm}
    \vspace{3mm}
    \resizebox{\textwidth}{!}{
        \begin{tabular}{l|cccccc|c}
            \toprule
            Method                               & Clipart                  & Infograph                & Painting                 & Quickdraw                & Real                     & Sketch                   & Average                  \\
            \midrule
            Local-Only                           & {24.27 (0.46)}           & {11.75 (0.36)}           & {23.97 (0.68)}           & {45.33 (0.75)}           & {56.61 (0.24)}           & {20.15 (0.36)}           & {37.64 (0.14)}           \\
            \texttt{FedAvg} \cite{fedavg}        & {54.06 (1.23)}           & {19.88 (0.86)}           & {44.72 (0.90)}           & {73.01 (0.92)}           & \underline{77.53 (0.39)} & \underline{45.21 (1.00)} & {60.36 (0.45)}           \\
            \texttt{FedProx} \cite{fedprox}      & {\bf 56.56 (1.36)}       & \underline{20.53 (1.17)} & \underline{44.87 (0.37)} & {65.56 (2.72)}           & {75.80 (0.73)}           & {45.17 (0.75)}           & {58.06 (0.72)}           \\
            \texttt{FedDyn} \cite{feddyn}        & {51.35 (4.20)}           & {19.26 (1.02)}           & {41.00 (2.35)}           & {71.56 (2.25)}           & {68.97 (1.80)}           & {43.84 (1.48)}           & {56.54 (1.14)}           \\
            \texttt{fedgen} \cite{fedgen}        & {55.42 (0.69)}           & {18.79 (0.97)}           & {44.17 (1.21)}           & \underline{75.72 (1.35)} & {76.87 (0.44)}           & {44.07 (0.99)}           & \underline{60.67 (0.62)} \\
            \texttt{FedBN} \cite{fedbn}          & {53.15 (0.38)}           & {19.27 (0.37)}           & {44.53 (0.45)}           & {73.51 (0.69)}           & {77.14 (0.62)}           & {44.45 (0.51)}           & {60.14 (0.31)}           \\
            \texttt{Per-FedAvg} \cite{perfedavg} & {41.10 (3.15)}           & {16.05 (0.54)}           & {34.70 (0.79)}           & {29.08 (2.58)}           & {66.13 (2.01)}           & {31.70 (0.89)}           & {40.56 (0.88)}           \\
            \texttt{FedPer} \cite{fedper}        & {44.06 (1.44)}           & {16.67 (0.70)}           & {37.48 (0.59)}           & {65.82 (3.01)}           & {72.48 (0.63)}           & {34.49 (0.83)}           & {53.48 (1.18)}           \\
            \texttt{FedRep} \cite{fedrep}        & {33.25 (0.86)}           & {14.62 (0.89)}           & {31.05 (0.97)}           & {56.84 (0.48)}           & {65.49 (0.84)}           & {26.16 (0.49)}           & {46.04 (0.18)}           \\
            \texttt{FedFomo} \cite{fedfomo}      & {42.13 (3.57)}           & {14.59 (0.72)}           & {31.71 (2.61)}           & {43.99 (1.31)}           & {65.16 (3.37)}           & {25.59 (2.07)}           & {43.04 (1.62)}           \\
            \texttt{f-FedAP} \cite{fedap}        & {53.20 (0.84)}           & {18.84 (0.80)}           & {44.42 (0.47)}           & {74.19 (0.76)}           & {77.30 (0.45)}           & {43.79 (0.42)}           & {60.22 (0.25)}           \\
            \midrule
            \texttt{pFedSim} (Ours)              & \underline{55.99 (1.17)} & {\bf 21.18 (0.69)}       & {\bf 46.88 (0.41)}       & {\bf 75.86 (0.45)}       & {\bf 78.20 (0.32)}       & {\bf 45.36 (0.86)}       & {\bf 61.90 (0.19)}       \\
            \bottomrule
        \end{tabular}
    }
    \caption{Average Accuracy over DomainNet with format mean(std). Bold and underline stand for the best and second-best results. The average results is the weighted average of accuracies of all domains.}
    \label{tb:domain_res}
\end{table}

\noindent{\bf Dataset Description}  DomainNet contains natural images coming from 6 different domains: Clipart, Infograph, Painting, Quickdraw, Real and Sketch. All domains contain data belonging to the same label space, \emph{i.e.}, 345 classes in total, but images look quite different to each other and have various sizes. DomainNet is widely used to evaluate the capability of domain generalization. In FL, the shift between domains can be considered as the feature non-IID nature \cite{fedbn}.

\noindent{\bf Experimental Settings} Due to the constraint of computation resources, we only sample 20 classes in DomainNet (345 classes in total) with 30\% data from each class for this experiment, and uniformly rescale images to $64 \times 64$. We list the data number of each domain in Table \ref{tb:domain}. Each domain (with 6 domains in total) is partitioned  into 20 FL clients (\emph{i.e.}, 120 FL clients in total). Each FL client contains data from only one domain. The data of FL clients belonging to the same domain are label IID but feature non-IID. We train and evaluate all aforementioned methods using MobileNetV2 \cite{mobilenetv2} under the same experiment setting. We report the average accuracy results by using 5 random seeds in Table \ref{tb:domain_res}.

\noindent{\bf Discussion} From Table \ref{tb:domain_res}, we can observe that: 1) In the feature non-IID scenario, traditional FL methods generally perform better than most pFL methods, except \texttt{pFedSim}, which is different from the most cases in Table \ref{tb:res}. It shows that the label non-IID nature can degrade traditional FL method performance more severely than the feature non-IID nature in FL, which is highly correlated with the classifier part. Due to the label non-IID nature, classifier parameters shift drastically and thus significantly weaken the global classifier, resulting in performance deterioration of a single global model. Under this setting, pFL methods (\emph{e.g.}, \texttt{FedPer} \cite{fedper}) that personalize classifiers perform better than traditional FL methods. However, the feature non-IID nature does not involve the non-IID of labels. Thus, we infer that the gain from personal classifiers is little. As a result, traditional FL baselines generally outperform pFL baselines that rely primarily on personalized classifiers. 2)  Our \texttt{pFedSim} method can guarantee performance superiority because it not only relies on personalization of the classifier, but also warming up (generalization phase) and the classifier similarity-based feature extractor aggregation. The result shows that our method \texttt{pFedSim} outperforms baselines under both label and feature non-IID settings, indicating  the robustness of our method.

\section{Discussion}
\subsection{Broader Impact}
In this work, we propose a novel pFL method by using model classifiers to evaluate client similarity. While the primary area studied in this paper (personalized FL) has a significant societal impact since FL solutions have been and are being deployed in many industrial systems.  Our work contributes to data privacy protection and improves the utility of the personalized model without revealing additional information other than model parameters. Thus, the impact of this work lies primarily in improving the utility of the personalized model while preserving data privacy. The broader impact discussion on this work is not applicable.

\subsection{Limitations}
We summarize the limitations of \texttt{pFedSim} as follows:
\begin{itemize}
    \item \texttt{pFedSim} relies heavily on the classifier part of a neural network, so \texttt{pFedSim} is not suitable for solving problems using models without a classifier, \emph{e.g.}, image segmentation and content generation.
    \item \texttt{pFedSim} is designed to solve problems in data non-IID scenarios.  When the degree of data non-IID is not significant, the improvement of \texttt{pFedSim} may be slight, which is the common dilemma confronted by most pFL methods.
    \item \texttt{pFedSim} requires the cloud server to store personalized models belonging to all FL clients, which requests additional storage resources on the cloud server side.
\end{itemize}

\begin{table}[htb!]
    \setlength{\abovecaptionskip}{5mm}
    \centering
    \resizebox{\textwidth}{!}{
        \begin{tabular}{llc}
            \toprule
            Component  & Layer                                                                      & Repetition                  \\
            \midrule
            \multirow{40}{*}{\shortstack{Feature                                                                                  \\Extractor}} & Conv2D($in$=3, $out$=32, $kernel$=3, $stride$=1, $pad$=1)                  & $\times1$                   \\
            \cmidrule(lr){2-3}
                       & Conv2D($in$=32, $out$=32, $kernel$=3, $stride$=1, $pad$=1, $groups$=32)    & \multirow{2}{*}{$\times 1$} \\
                       & Conv2D($in$=32, $out$=16, $kernel$=1, $stride$=1)                                                        \\
            \cmidrule(lr){2-3}
                       & Conv2D($in$=16, $out$=96, $kernel$=1, $stride$=1)                          & \multirow{3}{*}{$\times 1$} \\
                       & Conv2D($in$=96, $out$=96, $kernel$=3, $stride$=2, $pad$=1, $groups$=96)                                  \\
                       & Conv2D($in$=96, $out$=24, $kernel$=1, $stride$=1)                                                        \\
            \cmidrule(lr){2-3}
                       & Conv2D($in$=24, $out$=144, $kernel$=1, $stride$=1)                         & \multirow{3}{*}{$\times 1$} \\
                       & Conv2D($in$=144, $out$=144, $kernel$=3, $stride$=1, $pad$=1, $groups$=144)                               \\
                       & Conv2D($in$=144, $out$=24, $kernel$=1, $stride$=1)                                                       \\
            \cmidrule(lr){2-3}
                       & Conv2D($in$=24, $out$=144, $kernel$=1, $stride$=1)                         & \multirow{3}{*}{$\times 1$} \\
                       & Conv2D($in$=24, $out$=144, $kernel$=3, $stride$=2, $pad$=1, $groups$=144)                                \\
                       & Conv2D($in$=144, $out$=32, $kernel$=1, $stride$=1)                                                       \\
            \cmidrule(lr){2-3}
                       & Conv2D($in$=24, $out$=144, $kernel$=1, $stride$=1)                         & \multirow{3}{*}{$\times 2$} \\
                       & Conv2D($in$=32, $out$=192, $kernel$=3, $stride$=1, $pad$=1, $groups$=192)                                \\
                       & Conv2D($in$=192, $out$=32, $kernel$=1, $stride$=1)                                                       \\
            \cmidrule(lr){2-3}
                       & Conv2D($in$=32, $out$=192, $kernel$=1, $stride$=1)                         & \multirow{3}{*}{$\times 1$} \\
                       & Conv2D($in$=192, $out$=192, $kernel$=3, $stride$=2, $pad$=1, $groups$=192)                               \\
                       & Conv2D($in$=192, $out$=64, $kernel$=1, $stride$=1)                                                       \\
            \cmidrule(lr){2-3}
                       & Conv2D($in$=64, $out$=384, $kernel$=1, $stride$=1)                         & \multirow{3}{*}{$\times 3$} \\
                       & Conv2D($in$=384, $out$=384, $kernel$=3, $stride$=1, $pad$=1, $groups$=384)                               \\
                       & Conv2D($in$=384, $out$=64, $kernel$=1, $stride$=1)                                                       \\
            \cmidrule(lr){2-3}
                       & Conv2D($in$=64, $out$=384, $kernel$=1, $stride$=1)                         & \multirow{3}{*}{$\times 1$} \\
                       & Conv2D($in$=384, $out$=384, $kernel$=3, $stride$=1, $pad$=1, $groups$=384)                               \\
                       & Conv2D($in$=384, $out$=96, $kernel$=1, $stride$=1)                                                       \\
            \cmidrule(lr){2-3}
                       & Conv2D($in$=96, $out$=576, $kernel$=1, $stride$=1)                         & \multirow{3}{*}{$\times 2$} \\
                       & Conv2D($in$=576, $out$=576, $kernel$=3, $stride$=1, $pad$=1, $groups$=576)                               \\
                       & Conv2D($in$=576, $out$=96, $kernel$=1, $stride$=1)                                                       \\
            \cmidrule(lr){2-3}
                       & Conv2D($in$=96, $out$=576, $kernel$=1, $stride$=1)                         & \multirow{3}{*}{$\times 1$} \\
                       & Conv2D($in$=576, $out$=576, $kernel$=3, $stride$=2, $pad$=1, $groups$=576)                               \\
                       & Conv2D($in$=576, $out$=160, $kernel$=1, $stride$=1)                                                      \\
            \cmidrule(lr){2-3}
                       & Conv2D($in$=160, $out$=960, $kernel$=1, $stride$=1)                        & \multirow{3}{*}{$\times 2$} \\
                       & Conv2D($in$=960, $out$=960, $kernel$=3, $stride$=1, $pad$=1, $groups$=960)                               \\
                       & Conv2D($in$=960, $out$=160, $kernel$=1, $stride$=1)                                                      \\
            \cmidrule(lr){2-3}
                       & Conv2D($in$=160, $out$=960, $kernel$=1, $stride$=1)                        & \multirow{3}{*}{$\times 1$} \\
                       & Conv2D($in$=960, $out$=960, $kernel$=3, $stride$=1, $pad$=1, $groups$=960)                               \\
                       & Conv2D($in$=960, $out$=320, $kernel$=1, $stride$=1)                                                      \\
            \cmidrule(lr){2-3}
                       & Conv2D($in$=320, $out$=1280, $kernel$=1, $stride$=1)                       & \multirow{3}{*}{$\times 1$} \\
                       & AvgPool2D($out$=(1,1))                                                                                   \\
                       & Flatten()                                                                                                \\
            \midrule
            Classifier & FC(1280, $num\_classes$)                                                   & $\times1$                   \\
            \bottomrule
        \end{tabular}
    }
    \caption{MobileNetV2 Architecture. Conv2D consists of a 2D convolution layer, batch normalization layer and ReLU6 activation layer. $Groups$ property shown in Conv2D() means the depthwise convolution operation; AvgPool2D() adaptively pools input to match the output size request.}
    \label{tb:mobile_arch}

\end{table}